\documentclass{egpubl}
\usepackage{pg2026p}

\WsPoster          % uncomment for Poster contribution
\usepackage[T1]{fontenc}
\usepackage{dfadobe}  

\usepackage{cite}  % comment out for biblatex with backend=biber
\BibtexOrBiblatex
\electronicVersion
\PrintedOrElectronic
\ifpdf \usepackage[pdftex]{graphicx} \pdfcompresslevel=9
\else \usepackage[dvips]{graphicx} \fi

\usepackage{egweblnk}

\usepackage{xcolor}

\usepackage{booktabs}
\usepackage{multirow}
\title[Constrained Iterative Refinement for SVG Generation]%
{Evaluating Constrained Iterative Refinement for \\ Scalable Vector Graphics Generation with Off-the-Shelf VLMs}

\author[Perlman]{
\parbox{\textwidth}{
\centering
Matthew Perlman$^{1}$ \orcid{0009-0008-3848-5695},
James Beetham$^{2}$ \orcid{0009-0008-9602-2875},
Niels Da Vitoria Lobo$^{2}$ \orcid{0000-0001-5354-2805},
Amrit Singh Bedi$^{2}$ \orcid{0000-0002-8807-2695},
Mubarak Shah$^{2}$ \orcid{0000-0001-6172-5572}
}
\\
\parbox{\textwidth}{\centering
$^{1}$University of Massachusetts Amherst \\
$^{2}$University of Central Florida
}
}

\begin{document}

% \teaser{
%  \includegraphics[width=0.9\linewidth]{pipeline_labels.png}
%  \centering
%   \caption{Example images generated by GPT-5.6 using text prompt "A robot painting".}
% \label{fig:example_images}
% }

\maketitle

% ---------------------------------------------------------------------------

\begin{abstract}
Scalable Vector Graphics (SVGs) power much of the modern visual ecosystem, yet state-of-the-art generative models focus almost entirely on rasterized images. We explore whether inference-time methods can unlock SVG generation capabilities in off-the-shelf vision-language models (VLMs). We systematically evaluate a constrained iterative refinement harness that combines visual feedback, structured editing, and constrained decoding to characterize the capabilities and limitations of current VLMs for SVG generation. Across multiple VLMs and generation settings, we find that constrained decoding improves compilation success rates, while iterative refinement reveals a deficit in visual reasoning and self-correction. Our results highlight both the promise and current limitations of using inference-time methods to adapt general-purpose VLMs for SVG generation.




\end{abstract}

% ---------------------------------------------------------------------------

\section{Introduction}

% SVGs are widely used for web, interface, and illustration content because they are resolution-independent, compact, and directly editable. Despite the practical benefits of SVGs, generating them remains challenging for general-purpose VLMs: the model must jointly produce valid code, reason about spatial relationships, and translate precise visual details into a new medium. Existing SVG generation methods address these challenges through specialized architectures or task-specific training~\cite{omnisvg}, decoupling them from the rapidly improving ecosystem of general-purpose multimodal models.

SVGs are widely used for web, interface, and illustration content because they are resolution-independent, compact, and directly editable. Despite the practical benefits of SVGs, generating them remains challenging for general-purpose VLMs: the model must jointly produce valid code, reason about spatial relationships, and translate precise visual details into a new medium. Existing SVG generation methods~\cite{omnisvg} address these challenges through specialized architectures or task-specific training, decoupling them from the rapidly improving frontier multimodal models.

We investigate whether inference-time methods are sufficient to make off-the-shelf VLMs effective SVG generators. We introduce a lightweight harness: an external framework that wraps a VLM with rendering, visual feedback, structured editing, and constrained generation. Using this harness, we evaluate iterative refinement, four levels of editing structure, and grammar-constrained decoding ~\cite{geng2023grammar} for text-to-SVG and image-to-SVG generation across multiple model scales. Our study characterizes when these mechanisms improve syntactic reliability, affect visual quality, and how effectively VLMs translate rendered feedback into code changes.

% ---------------------------------------------------------------------------

\begin{figure}[h]
    \centering
    \includegraphics[width=\linewidth]{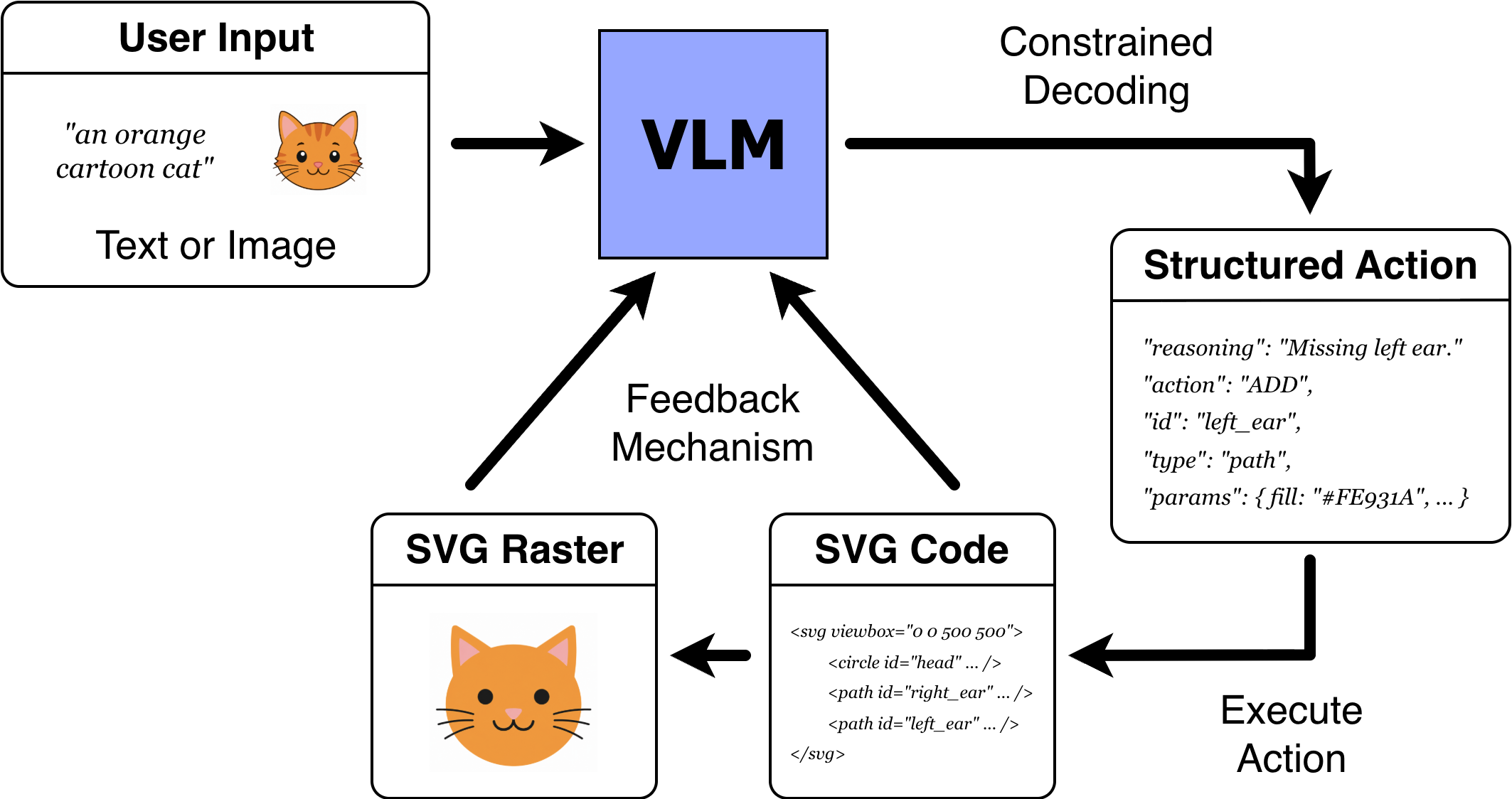}
    \caption{
    % Our Constrained Iterative Refinement Harness.
    % We equip off-the-shelf VLMs with two complementary, inference-time mechanisms: iterative refinement based on rendered visual feedback and grammar-constrained decoding that restricts outputs to appropriate SVG and JSON syntax. Together, these mechanisms target the weaknesses of general-purpose VLMs, allowing the model to iteratively refine and correct its output while remaining bounded to a syntactically valid generation space.
    % Our proposed harness. Given a user query, the VLM generates a \textbf{constrained output} specifying an edit to the current SVG. We parse the resulting action, update and render the SVG, and provide the updated code and image back to the VLM for the next \textbf{refinement iteration}.
    Our proposed harness. Given a user query, the VLM generates a \textbf{constrained output} specifying an SVG edit. We parse the action, update and render the SVG, and return the code and image to the VLM for the next \textbf{refinement iteration}.
    }
    \label{fig:system_diagram}
\end{figure}

% 9b and 27b macro avg
\begin{table*}[h!]
\centering
\footnotesize
\setlength{\tabcolsep}{8pt}
\renewcommand{\arraystretch}{0.9}
\begin{tabular}{llcccccccc}
\toprule
&
&
\multicolumn{4}{c}{\textbf{Text$\rightarrow$SVG}} &
\multicolumn{4}{c}{\textbf{Image$\rightarrow$SVG}} \\
\cmidrule(lr){3-6}
\cmidrule(l){7-10}
\textbf{Decoding} &
\textbf{Pipeline} &
\textbf{Compile$\uparrow$} &
\textbf{CLIP$\uparrow$} &
\textbf{Aesthetic.$\uparrow$} &
\textbf{\#Tokens} &
\textbf{Compile$\uparrow$} &
\textbf{DINO$\uparrow$} &
\textbf{LPIPS$\downarrow$} &
\textbf{\#Tokens} \\
\midrule
\multirow{4}{*}{Unconstrained}
& Single Turn      & \textbf{0.837} & \textbf{0.255} & \textbf{4.655} & 1.4k & \underline{0.630} & \textbf{0.704} & \underline{0.396} & 1.7k \\
& Unstructured     & 0.475 & \underline{0.246} & \underline{4.560}  & 4.8k & 0.113 & \textbf{0.704} & \textbf{0.372} & 10.7k \\
& Semi-Structured  & 0.180 & 0.205 & 4.415 & 7.9k & 0.500 & 0.617 & 0.418 & 8.8k \\
& Structured       & \underline{0.785} & 0.226 & 4.480  & 3.3k & \textbf{0.650} & 0.633 & 0.413 & 8.1k \\
\cmidrule(l){1-10}
\multirow{4}{*}{Constrained}
& Single Turn      & \textbf{0.830} & \underline{0.220} & \textbf{4.530}  & 0.6k  & \textbf{0.830} & 0.411 & 0.579 & 0.4k  \\
& Unstructured     & 0.599 & \textbf{0.242} & \underline{4.485} & 5.2k & 0.647 & \textbf{0.617} & \underline{0.430} & 5.1k \\
& Semi-Structured  & 0.512 & 0.206 & 4.445 & 2.7k & 0.375 & 0.464 & 0.506 & 3.2k \\
& Structured       & \underline{0.790} & 0.218 & 4.440  & 2.4k & \underline{0.695} & \underline{0.603} & \textbf{0.420} & 3.3k \\
\bottomrule
\end{tabular}
\caption{Results averaged across Qwen3.5 9B and 27B on MMSVG-Bench.}
\label{tab:benchmark_scores_macroavg_9b27b}
\end{table*}

\section{System Overview}

% We introduce a harness that equips off-the-shelf VLMs with two complementary inference-time mechanisms: iterative refinement using rendered visual feedback and constrained decoding over SVG and JSON outputs. Together, these mechanisms address two key limitations of general-purpose VLMs for SVG generation: difficulty correcting visual errors and reliably producing structured code. The harness enables iterative self-correction while keeping model outputs within a constrained, syntactically valid generation space.
% We introduce a harness combining iterative refinement with rendered visual feedback and constrained decoding over SVG and JSON outputs. These mechanisms target two key challenges: using visual feedback to correct errors and reliably expressing edits as valid structured code. Together, they enable iterative SVG revision without additional model training.
% Our harness, shown in Figure~\ref{fig:system_diagram}, combines iterative refinement using visual feedback with constrained decoding of VLM outputs, aiming to address the challenges of visual self-correction and generation.
% Our harness, shown in Figure~\ref{fig:system_diagram}, combines iterative refinement with visual feedback and the constrained decoding of VLM outputs, enabling visual self-correction and reliable structured generation.
Our harness, shown in Figure~\ref{fig:system_diagram}, combines constrained decoding of VLM outputs with iterative refinement using visual feedback, enabling reliable structured generation and visual self-correction.

% We introduce a harness (Figure~\ref{fig:system_diagram}).

\subsection{Constrained Decoding}

% Many smaller models struggle to produce correct JSON structure and SVG code. We explore various inference-time constraints to address this failure mode.

% For closed-source models, we constrain outputs based on JSON schemas, ensuring models adhere to the correct property names and structure.
% For open-source models, we further use a context-free grammar (CFG) over valid SVG structures. To ensure constrained decoding is efficient and tractable, we restrict the grammar to a core set of elements (e.g., \texttt{rect}, \texttt{circle}, \texttt{path}) and their respective attributes rather than enforcing the full SVG specification, producing syntactically valid SVGs by construction. Grammar-constrained decoding is \textit{not} exposed by closed-source models.

Many smaller models struggle to produce correct JSON structure and SVG code, motivating inference-time constraints. For closed-source models (GPT-5.6), we enforce JSON schemas to ensure valid property names and structure; for open-source models, we use a context-free grammar (CFG) over a tractable subset of SVG elements (e.g., \texttt{rect}, \texttt{circle}, \texttt{path}) and their attributes, producing syntactically valid SVGs by construction. Grammar-constrained decoding is \textit{not} exposed by most closed-source models.

% \subsection{Iterative Refinement}

% Instead of generating a complete SVG in a single step, we treat generation as a sequence of iterative edits. At each step, the model observes the original user instruction, current SVG code, and its rasterization, and produces an update to the scene.

\subsection{Iterative Refinement}

% We explore four refinement variants:
We explore three refinement variants which aim to improve the generated SVG at each iteration, along with a single-turn baseline:

\textbf{Single Turn.} A baseline where the model generates the full SVG once, with no iterative refinement or visual feedback.

\textbf{Unstructured Refinement.} The model regenerates the entire SVG at each iteration, rebuilding the image from scratch.

% for reference, initial draft:

% Semi-Structured Refinement. At each step, the model emits an
% action (Add, Update, Remove, or Exit) alongside a target element
% ID and, where relevant, an accompanying SVG code snippet.
% Structured Refinement. At each step, the model emits an action
% (Add, Update, Remove, or Exit) alongside a target element ID and,
% where relevant, a fully parameterized set of element attributes.

% \textbf{Semi-Structured Refinement.} The model selects an action each iteration and generates an SVG snippet used to add or replace a single target element.
\textbf{Semi-Structured Refinement.} The model selects an action and generates an SVG code snippet used to add or replace an element.

\textbf{Structured Refinement.} The model outputs an action and element parameters (e.g. a rectangle's position, width, height, and color) which the harness converts into a full SVG snippet.

% \textbf{Structured Refinement.} The model outputs element parameters, e.g. x, y, width and height of a rectangle, which the harness converts into the corresponding SVG snippet to add or replace an element.

% \textbf{Semi-Structured Refinement.} At each step, the model specifies an action, a target element ID when required, and an SVG snippet to instantiate or replace the target element.

% \textbf{Structured Refinement.} At each step, the model specifies an action alongside a target element ID when required, and a list of parameters the harness converts into an SVG snippet.

% Each variant outputs free-form reasoning text before its code/actions/parameters in JSON. Semi-Structured and Structured use an action space of \textit{Add}, \textit{Update}, \textit{Remove}, or \textit{Exit}. Unstructured refinement is capped at 10 steps, while the other variants allow up to 20 for more granular edits.
% All variants include free-form reasoning before the generated code or action in JSON. Unstructured refinement is capped at 10 steps, while the more granular Semi-Structured and Structured variants allow up to 20.
All variants include free-form reasoning before the generated code or action in JSON. Semi-Structured and Structured use the actions \textit{Add}, \textit{Update}, \textit{Remove}, and \textit{Exit}, with a target element ID when required. Unstructured refinement is capped at 10 steps, while the more granular Semi-Structured and Structured allow up to 20.

% ---------------------------------------------------------------------------

\section{Results}

\begin{figure}[t]
    \centering
    \includegraphics[width=\linewidth]{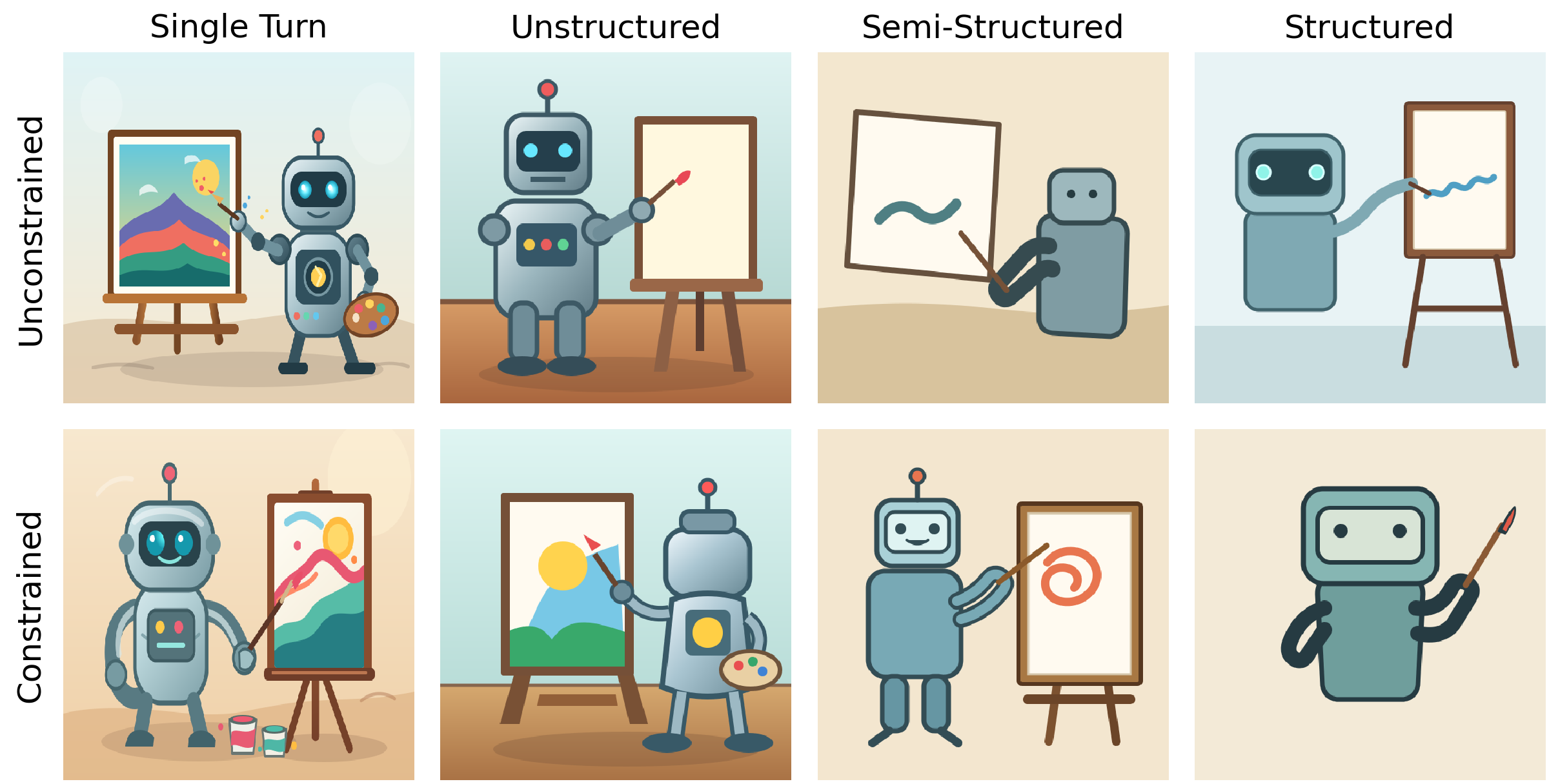}
    \caption{Images generated by GPT-5.6 using each pipeline and decoding strategy from the Text-to-SVG prompt "A robot painting".}
    \label{fig:example_images}
\end{figure}

We evaluate our framework on MMSVG-Bench~\cite{omnisvg} across multiple model scales
(Qwen3.5 9B and 27B)~\cite{qwen35}, with and without grammar-constrained decoding, and across all four generation pipelines (Table~\ref{tab:benchmark_scores_macroavg_9b27b}).
Metrics are computed over successfully compiled images.
We provide example images generated by GPT-5.6~\cite{gpt56} using each pipeline and decoding strategy in Figure~\ref{fig:example_images}.

\textbf{Constrained Decoding Hurts Visual Quality.} Across visual metrics and generation settings, constrained decoding generally resulted in lower quality images. We hypothesize that enforcing our limited grammar steers token generation into unlikely and visually flawed SVG elements.

\textbf{Constrained Decoding Can Improve Compilation.} We find that constrained decoding substantially improves compilation rates for most pipelines, a result consistent with our initial hypothesis.

\textbf{Small Models Struggle With Visual Feedback.} We find that for the models tested, iterative refinement has an inconsistent effect on visual quality, and often lowers compilation rates by introducing more opportunities to break.
Our testing suggests small models struggle to link visual inconsistencies to the underlying SVG code.

% \textbf{Visual Quality and Validity are Distinct Objectives.} TODO

% ---------------------------------------------------------------------------

\section{Discussion}
% \section{Future Work}

Our results suggest that reliable SVG generation requires more than syntactic constraints or repeated visual feedback. Constrained decoding can improve validity while reducing visual quality, while iterative refinement exposes a limitation in current VLMs' ability to translate rendered errors into precise vector-level corrections. These findings highlight visual quality and structural validity as distinct objectives that must be jointly addressed.

In the future, richer action spaces, alternative visual feedback mechanisms, stronger constrained decoding strategies, and prompt meta-optimization all have the possibility to improve generation quality and success rate.
Beyond extending our harness, a deeper mechanistic analysis of the model's failure to incorporate visual feedback may inform improvements to general-purpose VLM spatial-reasoning capabilities.

\bibliographystyle{eg-alpha-doi}
\bibliography{egbibsample}

@inproceedings{omnisvg,
  title={OmniSVG: A Unified Scalable Vector Graphics Generation Model},
  author={Yang, Yiying and Cheng, Wei and Chen, Sijin and Zeng, Xianfang and Yin, Fukun and Zhang, Jiaxu and Wang, Liao and Yu, Gang and Ma, Xingjun and Jiang, Yu-Gang},
  booktitle={Advances in Neural Information Processing Systems (NeurIPS)},
  year={2025}
}

@misc{qwen35,
  title={Qwen3.5},
  author={{Qwen Team}},
  year={2026},
  howpublished={\url{https://huggingface.co/Qwen}},
  note={Accessed: 2026-07-13}
}

@misc{gpt56,
  title        = {GPT-5.6},
  author       = {{OpenAI}},
  year         = {2026},
  howpublished = {OpenAI API},
  url          = {https://developers.openai.com/api/docs/models},
  note         = {Accessed: 2026-07-13}
}

@inproceedings{geng2023grammar,
  title={Grammar-Constrained Decoding for Structured {NLP} Tasks without Finetuning},
  author={Geng, Saibo and Josifoski, Martin and Peyrard, Maxime and West, Robert},
  booktitle={EMNLP},
  year={2023},
  pages={10932--10952}
}

\end{document}